\documentclass[letterpaper]{article} 
\usepackage{aaai2026} 
\usepackage{times}  
\usepackage{helvet}  
\usepackage{courier}  
\usepackage[hyphens]{url}  
\usepackage{graphicx} 
\usepackage{natbib}  
\usepackage{caption} 
\usepackage{algorithm}
\usepackage{algorithmic}
\usepackage{amsmath}
\usepackage{rotating}
\usepackage{booktabs}

\usepackage{newfloat}
\usepackage{listings}
\DeclareCaptionStyle{ruled}{labelfont=normalfont,labelsep=colon,strut=off} 
\floatstyle{ruled}
\newfloat{listing}{tb}{lst}{}
\floatname{listing}{Listing}
\title{CCD-Bench: Probing Cultural Conflict in Large Language Model Decision-Making}

\author{
    Hasibur Rahman, 
    Hanan Salam
}

\affiliations{
    New York University Abu Dhabi, Abu Dhabi, United Arab Emirates\\
    hasibur.rahman@nyu.edu, hanan.salam@nyu.edu
}

\usepackage{bibentry}

\begin{document}        
\maketitle

\begin{abstract}
Large language models (LLMs) increasingly shape interpersonal and societal decision-making, yet their ability to navigate explicit conflicts between legitimate cultural values remains underexplored. 
Existing benchmarks focus on cultural knowledge (CulturalBench), value inference (WorldValuesBench), or single-axis bias (CDEval), but none assess how LLMs adjudicate when multiple cultural frameworks directly clash.
We introduce CCD-Bench, a benchmark for evaluating LLM decision-making under cross-cultural value conflict. CCD-Bench contains 2,182 open-ended dilemmas across seven domains, each with ten anonymized response options aligned with the ten GLOBE cultural clusters spanning 62 societies. Using a Stratified Latin Square design, we evaluate 17 leading LLMs and find clear biases: 
models favor Nordic Europe (20.2\%) and Germanic Europe (12.4\%), while Eastern Europe and Middle East \& North Africa responses are least preferred ($\approx$5–6\%). Although 87.9\% of model rationales reference multiple cultural dimensions, this pluralism is shallow-dominated by Future and Performance Orientation, with limited attention to Assertiveness or Gender Egalitarianism ($<$3\%). Ordering effects are negligible, and model similarity clusters by developer lineage rather than geography. CCD-Bench shifts evaluation from bias detection to pluralistic reasoning, revealing that current LLMs express a Western-centric, consensus-oriented worldview even when confronted with equally valid, culturally diverse alternatives.

\end{abstract}

\begin{links}
    \link{Code}{https://github.com/smartlab-nyu/CCD-Bench}
\end{links}

\section{Introduction}

Large Language models (LLMs) are increasingly deployed as mediators in interpersonal and societal decision-making, including conflict resolution and moral judgment tasks \cite{tan_robots_2024, pires_how_2025, kocak_llms_2024}. As conversational interfaces integrate LLMs into everyday applications, millions of users now rely on them for socially situated guidance. Even a mundane query, such as ``How should I email my boss to request leave?'', requires not only factual accuracy but also culturally sensitive judgment that aligns with local norms, values and expectations.

Yet the global adoption of LLMs reveals a critical challenge: aligning model recommendations with the full spectrum of cultural value systems. A growing body of evidence shows that LLMs inherit salient cultural leanings from their training data \cite{wang_cdeval_2024, karinshak_llm-globe_2024, lu_cultural_2025, alkhamissi_investigating_2024}, leading users to receive responses that, while coherent, may reflect narrow or inconsistent cultural assumptions. For instance,  the question ``How should responsibilities be shared within a family?'' can evoke equally legitimate but divergent answers, from merit-based distribution to age-based hierarchy to gender egalitarianism, depending on cultural context. Although prompt-engineering strategies that foreground local contextual cues can mitigate such effects\cite{xu_self-pluralising_2025, zhong_cultural_2024}, a deeper issue remains:
\textit{"How do LLMs resolve situations in which legitimate cultural values, such as hierarchy versus egalitarianism, or group harmony versus personal initiative, directly conflict?"}.
These systematic preferences raise an important distinction between bias and cultural trend. Following \citet{alkhamissi_investigating_2024}, we refer to these patterns as cultural trends rather than biases, emphasizing that they do not necessarily entail harm but highlight limitations in addressing pluralistic value systems.

Existing leading benchmarks, such as MMLU \cite{hendrycks_measuring_2021} and ELO \cite{zheng_judging_2023}, prioritize factual accuracy and stylistic quality,  overlooking the cultural nuance and social intelligence required in real-world interactions. Early alignment paradigms similarly prioritize universal principles such as helpfulness, honesty, and harmlessness (the ``HHH'' criteria) \cite{askell_general_2021, ouyang_training_2022, bai_training_2022}, while sidestepping the pluralistic and culturally contingent nature of human values \cite{barbara_herrnstein_smith_contingencies_1991}. Yet, real decisions often hinge on reconciling legitimate but conflicting value systems, a challenge where the ``HHH'' framework offers little guidance. 
Existing cultural benchmarks \cite{wang_cdeval_2024, chiu_culturalbench_2024, zhao_worldvaluesbench_2024} capture model tendencies along individual cultural axes, but none examine how LLMs reason when culturally grounded values directly collide. This leaves a critical gap in evaluating LLM decision-making within pluralistic, cross-cultural contexts.

We introduce CCD-Bench (Culture-Conflict Decision Benchmark), a benchmark for evaluating how LLMs navigate value-plural dilemmas: cases where multiple cultural worldviews are equally defensible. It assesses models' ability to engage in pluralistic reasoning, recognizing and justifying choices across competing cultural values.

Our diagnostic suite evaluates the 17 widely used LLMs (excluding specialized reasoning models), embedding all culturally salient information directly in the prompt to isolate the LLMs' intrinsic decision-making tendencies. The benchmark comprises 2,182 dilemmas across seven domains: Arts, Education, Family, Wellness, Work, Science, and Lifestyle, adapted from \citet{wang_cdeval_2024}. Building on evidence that many LLMs already encode cultural frameworks \cite{karinshak_llm-globe_2024}, we generate ten anonymized candidate answers for each dilemma using a state-of-the-art reasoning model, followed by human verification. Each response corresponds to one of the ten GLOBE cultural clusters: Anglo, Latin Europe, Nordic Europe, Germanic Europe, Eastern Europe, Latin America, Sub-Saharan Africa, Middle East \& North Africa, Southern Asia, and Confucian Asia, representing 62 societies \cite{house_2004_culture}. Instead of locating models along isolated cultural dimensions, CCD-Bench embeds the cultural trade-off within each dilemma: the model must select and justify one of ten equally valid, culturally grounded responses. This design directly exposes how LLMs reason when confronted with competing cultural imperatives.

We advance three main contributions:

\begin{enumerate}
\item \textbf{Benchmark}: We introduce CCD-Bench, the first benchmark for evaluating LLMs on navigating complex cross-cultural value conflicts, comprising 2,182 dilemmas across seven domains, each paired with ten anonymized culturally-distinct responses representing the GLOBE cultural clusters.

\item \textbf{Methodology}: We develop a generation and verification pipeline for producing culturally authentic alternatives, combining multi-cluster prompting, similarity pruning, human annotation, and a Stratified Latin Square design to ensure a fair assessment of value pluralism.

\item \textbf{Analysis}: We evaluate 17 leading LLMs on CCD-Bench, revealing systematic patterns in how  models navigate cultural conflicts and the extent to which they default to specific cultural frameworks.
\end{enumerate}
\section{Related Work}

\textbf{Cultural Alignment in LLMs.} Compelling evidence shows LLMs encode distinct cultural orientations. For instance, prompting GPT-4 in English tends to produce analytic, independence-oriented reasoning, whereas Chinese prompts evoke holistic, interdependent values \cite{lu_cultural_2025}. Similarly, \textit{CDEval} finds GPT-4 skewed toward low power distance and high individualism, while Baichuan-13B and Qwen-7B exhibit stronger collectivist tendencies \cite{wang_cdeval_2024}. Yet, such binary, one-dimension-at-a-time evaluations oversimplify pluralism. \textit{LLM-GLOBE} advances this direction using open-ended prompts to uncover tonal and rhetorical differences across cultures \cite{karinshak_llm-globe_2024}, while \citet{alkhamissi_investigating_2024} improves alignment through ``Anthropological Prompting.'' However, these studies stop short of testing how models reason when legitimate cultural values directly conflict.


\textbf{Cultural Knowledge and bias benchmarks.}
Several Benchmarks assess cultural understanding or bias in LLMs but similarly avoid explicit value conflict.
For instance, \textit{CulturalBench} \cite{chiu_culturalbench_2024} features 1,227 human-written questions across 45 regions and 17 topics to assess cultural knowledge, with variants for different difficulty levels. However, it focuses on recall and ambiguity handling rather than decision-making amid conflicting norms. \textit{WorldValuesBench} \cite{zhao_worldvaluesbench_2024} draws from the World Values Survey \footnote{https://www.worldvaluessurvey.org/wvs.jsp} to create millions of examples for predicting demographic-conditioned values, evaluating alignment via distribution metrics, but emphasizes prediction over resolution of conflicting perspectives. \textit{BLEnD} \cite{myung_blend_2024} tests everyday cultural knowledge with 52,600 QA pairs across 16 countries and 13 languages, revealing performance disparities for low-resource cultures, although it targets factual recall rather than value trade-offs. CAMeL \cite{naous_having_2024} and its extensions detect biases in associations, particularly between Arab and Western entities, using fill-in-the-blank prompts and parallel data, but center on implicit biases instead of explicit dilemmas requiring justification. These benchmarks illuminate cultural gaps but center on representation and factual recall rather than decision-making under normative conflict.

Despite significant progress, no existing work systematically examines how LLMs adjudicate cultural value conflicts: situations where multiple legitimate worldviews demand incompatible choices. Addressing this gap, CCD-Bench focuses on pluralistic dilemmas that require explicit reasoning about competing cultural imperatives.

\section{Methodology}
This section presents the methodology underlying CCD-Bench, a benchmark designed to evaluate how LLMs navigate culturally grounded value conflicts. We used GLOBE as it aligns with our dilemma structure, provides validated cultural value information for 62 societies, and distinguishes between practices and values \cite{house_2004_culture}. The 10-cluster granularity is sufficient for demographic analyses. The benchmark construction process follows a three-module pipeline encompassing Generation, Verification, and Evaluation. The Generation Module produces culturally diverse dilemmas and candidate responses aligned with the GLOBE framework; the Verification Module ensures authenticity and cross-cultural validity through human annotation and quality control; and the Evaluation Module standardizes presentation and model assessment procedures. Together, these modules establish a replicable framework for systematically examining LLM decision-making under cultural pluralism.

\subsection{Generation Module}
The Generation Module constructs the core of CCD-Bench, producing a diverse set of 2,182 open-ended cultural dilemmas, each paired with ten culturally grounded response alternatives. We build on the validated question corpus from CDEval \cite{wang_cdeval_2024}, the most comprehensive collection of culturally relevant scenarios to date, to ensure that CCD-Bench captures authentic cultural decision points while remaining comparable with prior cultural-AI evaluations. Starting from 2,953 short-form questions, we retain 2,204 open-ended items likely to elicit divergent advice across cultural contexts, excluding binary or factual questions. Each retained scenario is categorized into one of seven domains: Arts, Education, Family, Wellness, Work, Science, and Lifestyle, to represent everyday value conflicts.


Each dilemma is designed to contrast two or more competing cultural imperatives (e.g., hierarchy vs. egalitarianism, collectivism vs. individual autonomy). For each dilemma~$q$, we generate a set of $10$ \textit{culturally distinctive answers} $\{a_c\}_{c=1}^{10}$, each corresponding to one of the ten GLOBE clusters $\mathcal{C}=\{$Anglo, Latin~Europe, \dots, Confucian~Asia$\}$ \cite{house_2004_culture}.  To produce these alternatives, we employ a multi-cluster prompting strategy grounded in the GLOBE framework, instructing a state-of-the-art reasoning model (OpenAI~\texttt{o3-2025-04-16}) to generate one response per cluster in a single call. Each prompt embeds fenced \texttt{<cultural identity>} directives that encapsulate anonymized GLOBE descriptors. 
 The model is required to return: (i) a \textit{one–sentence answer} that omits explicit cultural self-reference and (ii) a \textit{one–sentence rationale} that references the corresponding GLOBE cultural dimensions.
Following \citet{balepur_which_2025},  rationales are included to enhance internal coherence between cultural identity and decision justification. 
A detailed prompt structure is provided on GitHub.

\paragraph{Similarity pruning and regeneration.} 
Although the multi-cluster prompting procedure should yield culturally diverse responses, we observed occasional lexical redundancy when answers from different clusters conveyed similar advice using overlapping phrasing. To mitigate this, we computed \textit{pairwise MinHash Jaccard similarity} \cite{cann_using_2025} over 3-gram shingles for each ten-tuple of responses. Any pair exceeding a similarity threshold of $\tau\!>\!0.70$ triggered regeneration of the more redundant answer, with a maximum of two attempts. The threshold was determined empirically: manual inspection confirmed that pairs above 0.70 exhibited minimal semantic differentiation, whereas pairs below retained meaningful cultural distinctions.
After pruning, 22 questions were removed, yielding 2,182 final dilemmas. Across these, the pairwise MinHash Jaccard similarity among the ten alternatives had a mean of $M=0.123$ with a standard deviation of $\sigma=0.109$ (N = \(\binom{10}{2}\times2182\) comparisons), confirming that regeneration effectively reduced near-duplicate content while preserving linguistic and semantic diversity. This process enhanced the discriminative validity of CCD-Bench by ensuring that each response remains lexically distinct yet culturally coherent, a result further verified through subsequent human evaluation.

\paragraph{Option ordering.} \citet{pezeshkpour_large_2024} demonstrated that the order of answer options in multiple-choice settings can significantly affect LLM  performance. To neutralize such serial-position effects, we employ a $10\times10$ \textit{Stratified Latin Square}, a domain-specific adaptation of the standard Latin square approach.  Independent Latin squares are generated for each of the seven domains to ensure balanced option ordering within domains, rather than globally across the corpus. Each cultural alternative appears exactly once in every ordinal position (1–10) and precedes or follows every other alternative exactly once, thereby satisfying both row and column orthogonality. The complete evaluation-prompt specification is available on GitHub to support transparency and reproducibility.

\subsection{Verification Module}
A team of five reviewers and one meta-reviewer confirmed that (i) each generated option would be viewed as \textit{reasonable} within the target cluster, (ii) did not mention the cultural cluster directly, and (iii) was semantically distinct from the other nine options. Non-conforming options are regenerated until all criteria are met, with a maximum of two attempts. None of the earlier 2,182 dilemmas were removed at this step. More details can be found on GitHub.

\subsection{Evaluation Module}
\label{subsec:analysis}
Our analytic pipeline decomposes evaluation logs into three layers: descriptive statistics, null-hypothesis significance tests, and robust similarity/diversity indices.

\paragraph{Descriptive layer.}
For each evaluation file, we record (i) absolute frequency $f_c$ of each cultural cluster $c \in \{1,\dots,10\}$, (ii) display position distribution $\pi = (\pi_1,\dots,\pi_{10})$, and (iii) domain-specific cluster counts. Raw tallies are used without smoothing.

\paragraph{Position-bias test.}
Under null hypothesis $H_0$ of equal likelihood for $k=10$ positions, let $\mathbf{O}$ be observed counts and $\mathbf{E} = (N/k, \dots, N/k)$ expected counts, with $N = \sum_i O_i$. The $\chi^2$ goodness-of-fit test yields
\begin{equation}
\chi^2 = \sum_{i=1}^k \frac{(O_i - E_i)^2}{E_i}, \qquad V = \sqrt{\frac{\chi^2}{N(k-1)}}, 
\end{equation}

where Cramér's $V$ follows thresholds: negligible $<0.10$, small $<0.30$, medium $<0.50$, large $\ge 0.50$ \cite{harald_cramer_mathematical_1999}.

\paragraph{Domain--cluster association.}
We construct a \( D \times C \) contingency table \( \mathbf{T} = [T_{ij}] \) for domains (\( D \)) vs. clusters (\( C \)). Pearson's \( \chi^2(\mathbf{T}) \) with \( (D - 1)(C - 1) \) degrees of freedom is reported with standardized residuals \( r_{ij} = \frac{T_{ij} - E_{ij}}{\sqrt{E_{ij}}} \) to identify over/under-represented pairs \cite{agresti_introduction_2019}.

\paragraph{Latin-square balance.}
Option orderings use a $10 \times 10$ Latin square \cite{bradley_complete_1958}. Let $A$ be the $10 \times 10$ matrix of cluster appearances per position across $Q$ questions. Imbalance is measured as
\begin{equation}
\Delta = \bigl\| A - (Q/10) \mathbf{1} \bigr\|_1, \qquad \mathrm{BS} = 100 \left( 1 - \frac{\Delta}{1.8 \cdot 10Q} \right)\%,
\end{equation}
with 1.8 as the worst-case normalization factor.

\paragraph{Diversity metrics.}

For cluster frequencies $\mathbf{p}$ and sorted counts $\mathbf{x}$ with $x_{(1)} \le \dots \le x_{(n)}$,

\begin{equation}
H(\mathbf{p}) = -\sum_c p_c \log_2 p_c,
\end{equation}
\begin{equation}
G(\mathbf{x}) = \frac{1}{n} \Bigl( n+1 - \frac{2}{\sum_i x_{(i)}} \sum_{i=1}^n (n+1-i) x_{(i)} \Bigr)
\end{equation}
\cite{shannon_mathematical_1949, dorfman_formula_1979}.
For each statistic $\widehat{\theta}$, we draw $B=5,000$ bootstrap replicates and report [2.5, 97.5] percentiles as 95\% confidence bands \cite{efron_introduction_1998}.

\begin{figure}[t]
\centering
\includegraphics[width=\columnwidth]{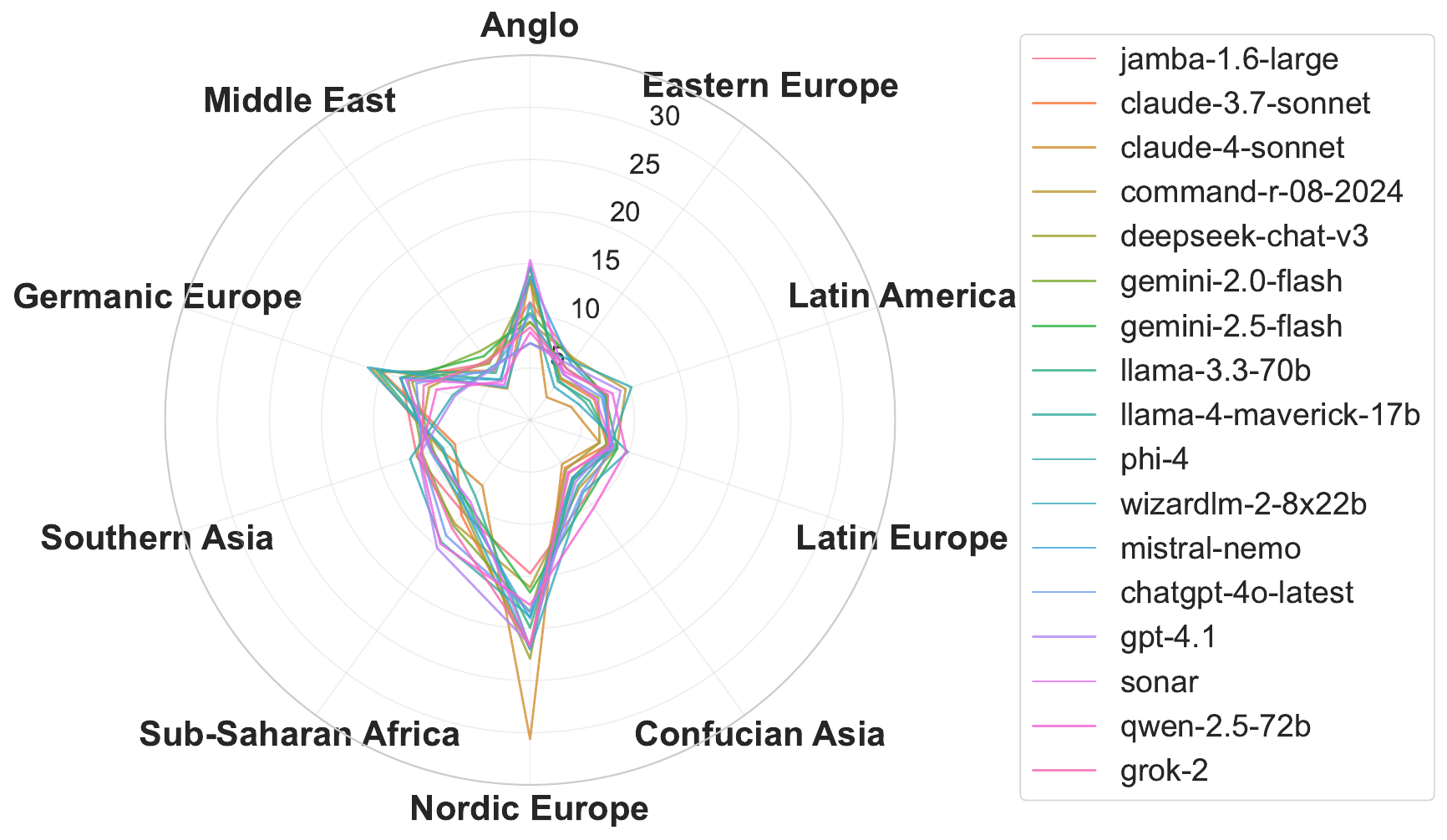}
\caption{Percentage distribution of cultural cluster selections across 17 LLMs. Each line represents one model, values normalized to percentages (0–100\%) for each of the 10 cultural clusters.}
\label{fig1}
\end{figure}

\paragraph{Model similarity.}
For LLMs with distributions $P$ and $Q$, symmetrized Kullback–Leibler divergence \cite{kullback_information_1997} is

\begin{equation}
D_{\mathrm{sym}}(P,Q) = \tfrac{1}{2} \sum_c \left[ P_c \log \frac{P_c}{Q_c} + Q_c \log \frac{Q_c}{P_c} \right],
\end{equation}
adding $10^{-10}$ to avoid $\log 0$. The matrix feeds average-linkage hierarchical clustering \cite{sneath_numerical_1973}.

\paragraph{Cluster–position bias.}
Repeating position-bias per cluster gives Cramér's $V_c$; mean $\overline{V} = \frac{1}{10} \sum_c V_c$ is the model-bias score.

\paragraph{Pluralism in rationales.}
Let \( I_i = 1 \) if rationale \( i \) cites \(\ge 2\) GLOBE dimensions (\( I_i = 0 \) otherwise), for \( i = 1, \dots, n \). The pluralism rate is \( R_{\text{plural}} = \frac{1}{n} \sum_i I_i \), reported per LLM and ensemble to assess value-reasoning breadth.

\section{Experiments}

We experiment with the 17 most popular LLMs for non-technical tasks, as ranked by OpenRouter \footnote{https://openrouter.ai/rankings}. Our inquiry is structured around five questions: 

\subsection{When offered culturally valid options, do LLMs still prefer certain clusters or values?(RQ1)}

\paragraph{A pronounced Nordic trend.}
Every single model in our sample selects the \textbf{Nordic Europe} option most often. The share ranges from $16.1\%$ (Cohere \texttt{command-r}) to a remarkable $30.6\%$ (Anthropic \texttt{Claude-4 Sonnet}), with an across-model mean of $M=20.2\%$. Put differently, the median model chooses Nordic-style advice roughly $\times2$ more often than chance, signaling an entrenched preference for highly egalitarian, consensus-driven solutions. Figure \ref{fig1} visualizes the raw selection frequencies for all $17$ LLMs.

\paragraph{Secondary leanings.}
Two clusters emerge as consistent runners-up: \textbf{Germanic Europe} ($M=12.4\%$): 7 LLMs list it as their second-most frequent choice. And \textbf{Sub-Saharan Africa} ($M=11.5\%$): likewise second for seven LLMs, underscoring that collectivist but low-power-distance reasoning resonates with many systems. A smaller subset (three LLMs, all OpenAI or Google) places the \textbf{Anglo} cluster in second position, yet its average share ($11.3\%$) still lags behind the Nordic and Germanic means.

\paragraph{Under-represented clusters.}
At the lower end of the spectrum lie \textbf{Eastern Europe} ($5.6\%$) and the \textbf{Middle East \& North Africa} ($5.8\%$).  No LLM exceeds an $8.2\%$ selection rate for either group, and the minimum for Eastern Europe drops to just $2.7\%$ in Anthropic \texttt{Claude-4 Sonnet}.  These figures suggest that narratives foregrounding high power distance or strong in-group collectivism are comparatively neglected.

\paragraph{Between-model variability.}
Cluster preferences vary in magnitude, but not in direction. The coefficient of variation is smallest for \textbf{Latin Europe} ($\text{CV}=0.09$), indicating that LLMs converge on a modest but stable share for this cluster. In contrast, \textbf{Anglo} selections display the greatest relative spread ($\text{CV}=0.23$), hinting that ``mainstream Western'' advice is more LLM-specific than often assumed.

\begin{figure}[t]
\centering
\includegraphics[width=\columnwidth]{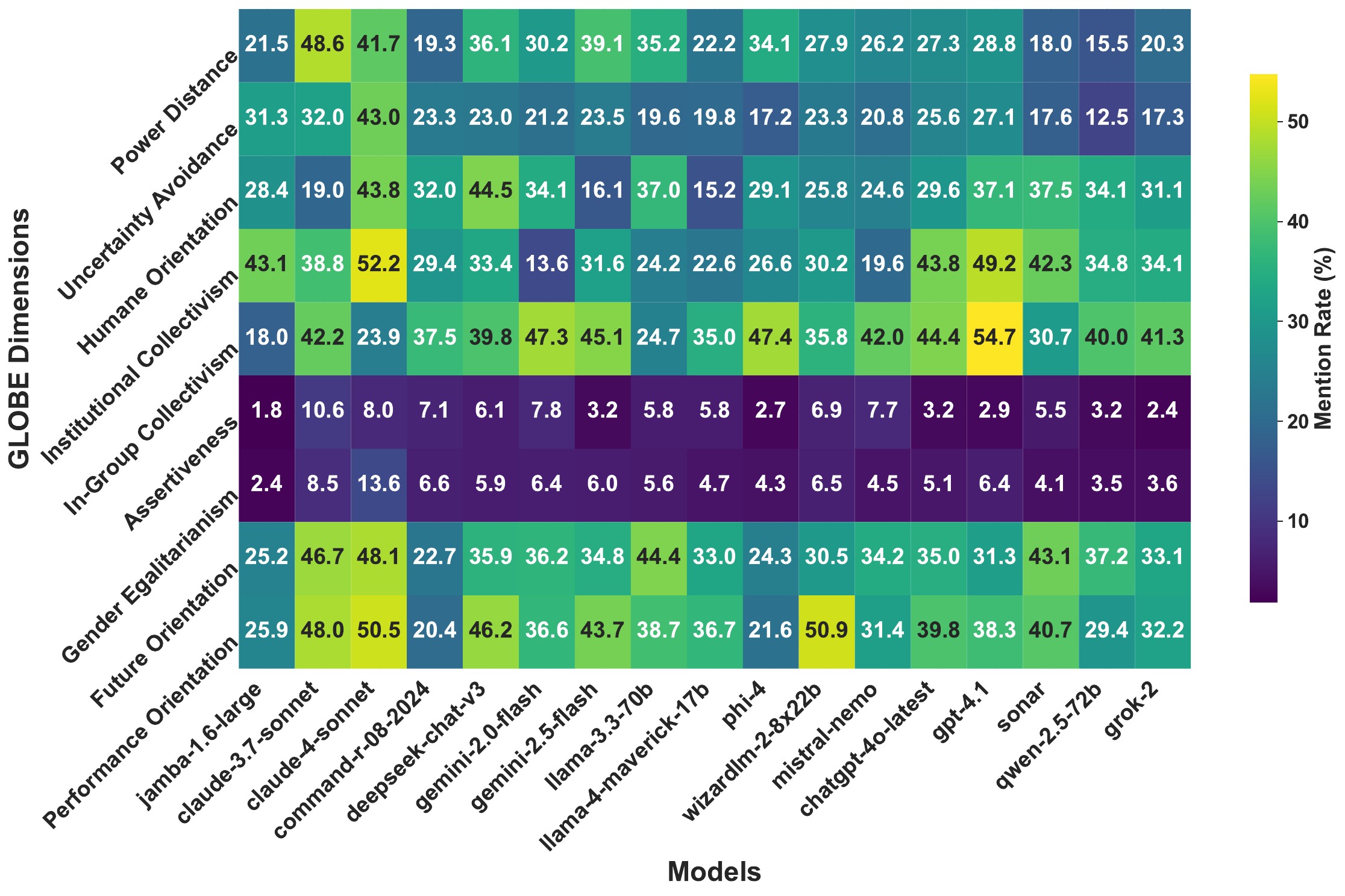}
\caption{Model-wise comparison of mention rates for nine GLOBE cultural dimensions cited in the rationales across 17 LLMs.}
\label{Globe_dimension}
\end{figure}

\paragraph{Which values do the LLMs foreground?}
Figure \ref{Globe_dimension} shows how often each system cites the nine GLOBE dimensions when justifying its selections. Two broad patterns emerge.

\begin{enumerate}
\item \textbf{A future-oriented, high-achievement consensus.} \\
Across the 17 LLMs, four dimensions dominate the rationales:

\begin{table}[H]
\centering
\small
\begin{tabular}{lcc}
\toprule
\textbf{Dimension} & \textbf{Mean Mention (\%)} & \textbf{s.d.} \\
\midrule
\textbf{In-Group Collectivism} & 38.2 & 9.5 \\
\textbf{Performance Orientation} & 37.1 & 9.4 \\
\textbf{Future Orientation} & 35.0 & 7.4 \\
\textbf{Institutional Collectivism} & 33.5 & 10.5 \\
\bottomrule
\end{tabular}
\caption{The four most cited GLOBE dimensions in the rationales of 17 LLMs.}
\label{tab:dimension_mentions}
\end{table}

All four sit well above the chance baseline of 11\% (one-ninth), indicating that, irrespective of the cultural cluster they ultimately pick, LLMs systematically justify their advice in terms of group solidarity, long-range thinking, and getting things done.

\item \textbf{Low appetite for confrontation or gender talk.}
At the opposite extreme, two dimensions are almost absent: \textit{Assertiveness: $5.2\%$ $(\sigma = 2.6)$} and \textit{Gender Egalitarianism: $5.7\%$ $(\sigma = 2.5)$}. No LLM mentions assertiveness in more than 11\% of its rationales, and the highest rate for gender egalitarianism is 13.6\% (Anthropic \texttt{Claude-4 Sonnet}). Even LLMs that frequently invoke \textit{Power Distance} (e.g.\ \texttt{Claude-3 Sonnet}, 48.6\%) do so via ``respect for hierarchy" rather than through a direct appeal to forceful negotiation.

\end{enumerate}

\paragraph{Between-model contrasts.}
While the aggregate rankings are stable, individual profiles diverge:
(i) \textbf{Anthropic LLMs} couple very high \textit{Performance Orientation} $\approx 50\%$ with the steepest \textit{Power Distance} scores, mirroring their Nordic cluster preference for meritocratic—but-still-hierarchical solutions. (ii) \textbf{Google’s Gemini 2.x} variants under-index on \textit{Institutional Collectivism} (13.6–31.6\%) yet over-index on \textit{Humane Orientation}, suggesting a more person-centred rationale style. (iii) \textbf{OpenAI’s GPT-4.1} is an outlier for \textit{In-Group Collectivism} (54.7\%), framing advice in terms of familial or team loyalty more than any other system.

Convergent evidence across clusters and evaluative dimensions yields a single, coherent conclusion. When confronted with ten anonymized, order-balanced alternatives, each of the 17 LLMs shows a marked preference for the normative frameworks of Northern and Western Europe, especially the Nordic and Germanic clusters, signaling an alignment imprint rather than stochastic variation. Their accompanying rationales tell the same story: although explicitly invited to invoke any of GLOBE’s nine cultural dimensions, LLMs repeatedly draw on future orientation, collective responsibility, and performance achievement, while rarely appealing to assertiveness or gender egalitarianism. Taken together, the results suggest prevailing alignment pipelines embed a conciliatory, progress-centred worldview emblematic of affluent liberal democracies, functional in many settings yet inadequate to capture humanity’s full breadth.

\subsection{Are these preferences domain-specific (e.g., religion vs.\ business) or globally consistent? (RQ2)}

To probe whether the cultural leanings uncovered in RQ1 persist across topical contexts, we split the 2,182 dilemmas into the seven CCD-Bench domains.

\paragraph{Dominant Nordic trend.}
In \textit{every} domain, the \textbf{Nordic Europe} option remains the single most frequent choice. Its mean share varies from a high of 23.9\% in \textsc{Family} to a low of 15.6\% in \textsc{Lifestyle}, but the ranking never drops below first place. Hence, the headline bias identified in RQ1 is not an artefact of a particular topic, such as business or science; it is a cross-domain constant of the evaluated LLMs.

\begin{center}
\small
\begin{tabular}{l}
\textbf{Domain:} Sci., Work, Educ., Wellness, Lifestyle, Family, Arts \\
\textbf{Nordic share (\%):} 21.9, 20.6, 21.2, 21.0, 15.6, 23.9, 15.8 \\
\textbf{Rank} 1, 1, 1, 1, 1, 1, 1
\end{tabular}
\end{center}

\paragraph{Which clusters fluctuate?}
While the top slot remains stable, the lower tail shifts according to context. The \textbf{least} referenced cluster varies by domain: \textit{Eastern Europe} in \textsc{Science}, \textit{Middle East \& North Africa} in \textsc{Work} and \textsc{Education}, \textit{Latin America} in \textsc{Wellness}, \textit{Latin Europe} in \textsc{Lifestyle}, and \textit{Confucian Asia} in \textsc{Arts}. Domain content thus nudges \textit{which} minority perspective is discounted. \textbf{Germanic Europe} and \textbf{Sub-Saharan Africa} exhibit the widest domain variance (s.d.\ $\approx$ 3 pp), rising in \textsc{Work} (Germanic) or \textsc{Wellness} (Sub-Saharan), yet receding elsewhere.

\paragraph{Statistical test.}
Collapsing all LLMs into a single \( 7 \times 10 \) contingency matrix and applying Pearson’s \( \chi^2 \) test yields \( \chi^2 = 256.7 \), \( df = 54 \), \( p < .001 \). However, the effect size is modest: Cramér’s \( V = .12 \) (95\% CI [.10, .14]), labeled ``small." In other words, the domain does influence cluster selection, but the global Nordic skew significantly outweighs its magnitude.

Cultural preferences in current LLMs are \textit{globally consistent}. Topical context shifts the edges of the distribution, deciding which clusters are least represented. Still, it does not perturb the overarching hierarchy: Nordic-style, low-power-distance, achievement-oriented advice dominates whether the user asks about science policy, workplace etiquette, family dilemmas, or the arts. The trends diagnosed in RQ1 are therefore systemic, not situational, and should be addressed at the alignment or data-curation level rather than via domain-specific patching.


\subsection{Do LLMs integrate multiple cultural dimensions or default to a single axis? (RQ3)}

\paragraph{High pluralism.}
Across the whole corpus ($N = 37,092$ rationales), 87.9\% cite two or more dimensions, indicating that most LLMs \textit{can} articulate multi-faceted justifications when prompted. Yet, the model spread is broad, ranging from a low of 68.6\% (Meta-Llama~4 Maverick) to a peak of 98.6\% (Anthropic Claude-4 Sonnet). 7 of the 17 systems exceed the 90\% mark, whereas three systems remain below 80\%. High pluralism is therefore common.

\paragraph{Pluralism $\,{\neq}\,$ cultural breadth.}
One might expect an LLM that distributes its selections broadly (i.e., exhibits high entropy) also to draw on more dimensions, but the data suggest otherwise. The Pearson correlation between entropy and pluralism is mildly \textit{negative} ($r = -0.22$). Cohere \texttt{command-r} exhibits the \textit{highest} entropy (3.28) yet one of the lower pluralism scores (78.5\%), while Anthropic Claude-4 shows the reverse. Integrating multiple dimensions and sampling widely across clusters are therefore distinct.

\paragraph{Which pairings dominate?}
Two results stand out: first, a single pairing explains one-fifth of all plural rationales, with \textit{Future Orientation (FO) + Performance Orientation (PO)} appearing 6,732 times (20.6\% of all plural explanations), no other pairing reaches 16\%, meaning LLMs overwhelmingly justify ``multi-dimensional'' decisions as ``long-term planning to achieve goals''; second, collectivist modifiers dominate the next tier, splicing \textit{Institutional Collectivism (IC)}, \textit{In-Group Collectivism (IGC)}, or \textit{Humane Orientation (HO)} onto either PO or FO, with the top six pairings, FO + PO (20.6\%), HO + IGC (15.3\%), IC + PO (14.4\%), FO + IC (12.9\%), IGC + Power Distance (PD) (12.4\%), and HO + IC (10.7\%), accounting for 86\% of all plural rationales. \paragraph{Dimension-level centrality.}
Summing across all pairings shows the frequency with which each GLOBE dimension enters multi-dimensional rationales: PO (16.2\%), FO (15.1\%), IC (14.9\%), IGC (13.5\%), HO (13.0\%), PD (12.5\%), Uncertainty Avoidance (UA) (9.5\%), Gender Egalitarianism (GE) (2.7\%), and Assertiveness (AS) (2.6\%). The first six account for over 85\% of mentions, with AS and GE remaining fringe ($<$3\% each).

\paragraph{Model signatures.}
The expanded per-model breakdown shows that: (i) \textbf{Anthropic models} over-index on the ``achievement + hierarchy'' tripod: PO, IC, and PD.
(ii) \textbf{Google Gemini 2.x} variants distinguish themselves through frequent IGC + PO pairings, signaling a ``team success'' rhetoric. (iii) \textbf{Meta-Llama 4 Maverick} is the only model where FO + PO accounts for less than 24\% of its plural rationales, yet it also exhibits the lowest pluralism rate overall.

LLMs integrate multiple cultural dimensions, but combinations are imbalanced: four mainstream values, FO, PO, and the two forms of Collectivism, dominate $\approx 90\%$ of multi-dimensional rationales, while AS and GE remain marginal. Thus, models exhibit limited pluralism, relying on a narrow subset of harmonious dimensions.

\subsection{Do Models Exhibit Systematic Ordering Bias When Selecting Among Options? (RQ4)}

\begin{figure}[h]
\centering
\includegraphics[width=\columnwidth]{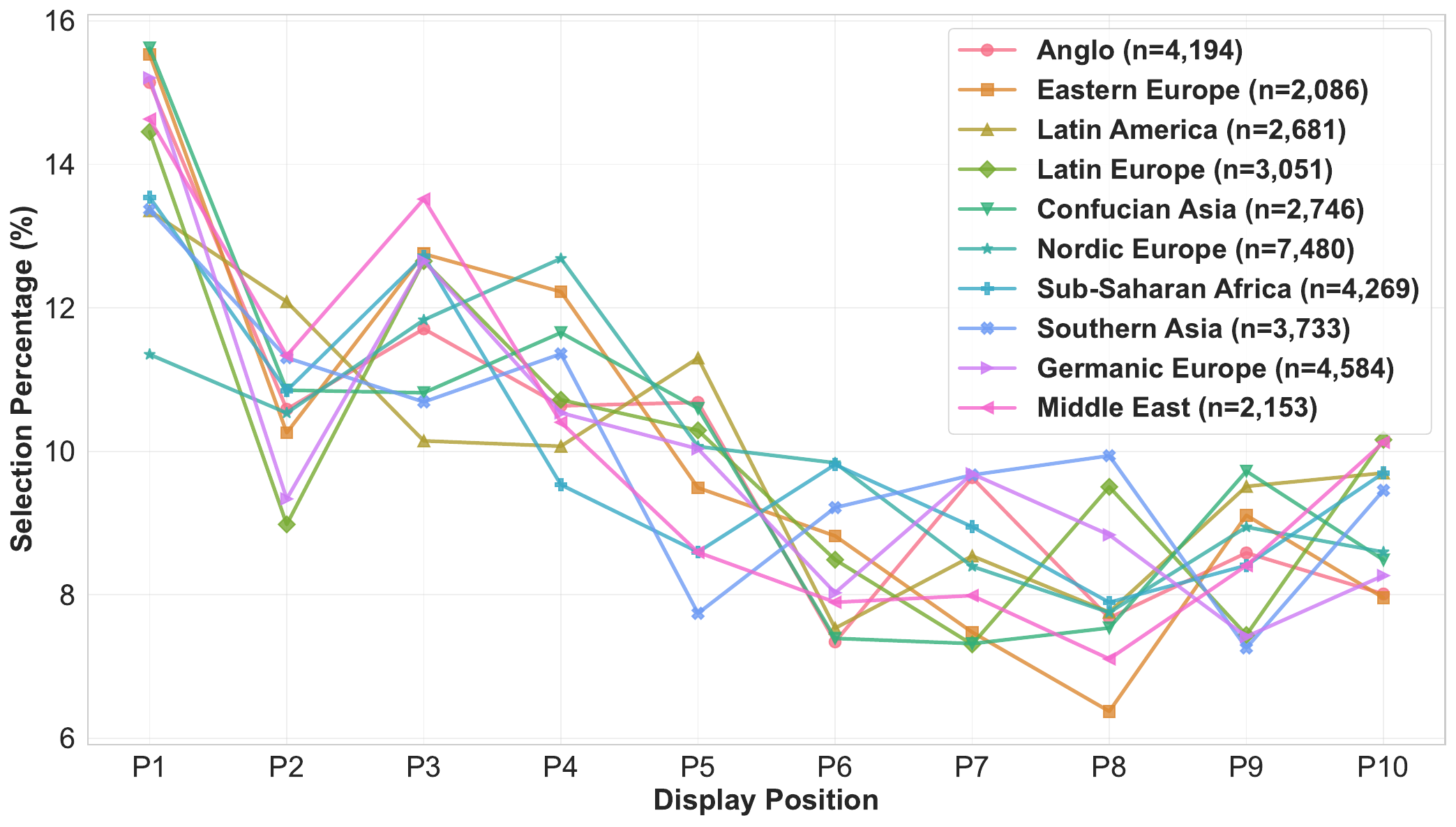}
\caption{Selection percentage of each cultural cluster across display positions (P1–P10), aggregated over all 17 LLMs.}
\label{fig:cluster_position_bias}
\end{figure}

\begin{figure*}[t]
  \centering
  \makebox[\textwidth][c]{%
    \includegraphics[height=0.394\textheight,width=\textwidth]{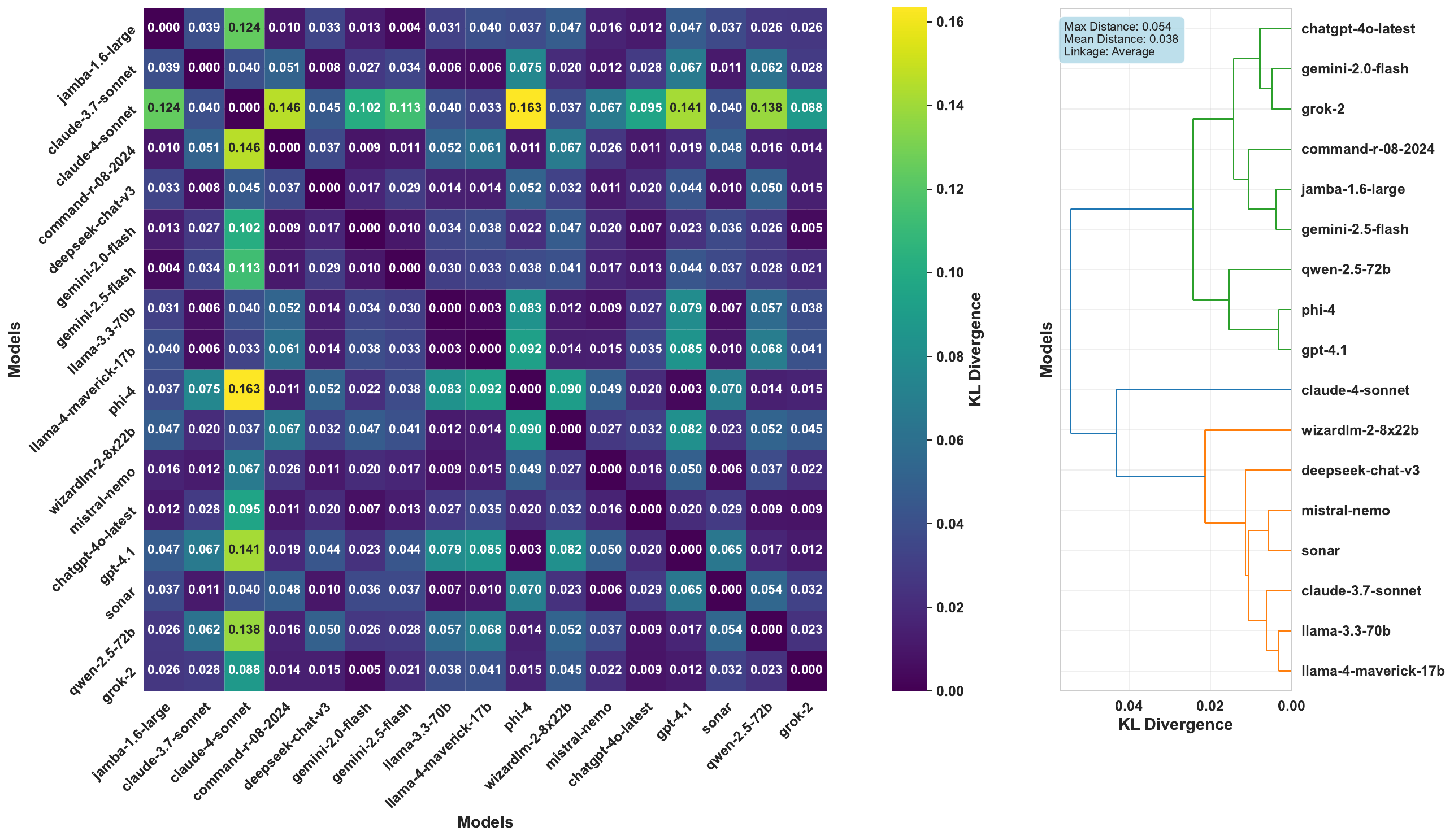}%
  }
  \caption{Heatmap of symmetrized KL-divergence between models' cultural cluster selections (left), with average-linkage hierarchical clustering dendrogram (right).}
  \label{fig:kl_divergence_heatmap}
\end{figure*}

Our generated corpus has a Latin-square balance score of $\approx 99.8\%$. Despite employing the Stratified Latin Square design to neutralize serial and residual ordering biases, these biases may persist if models exhibit systematic preferences for certain positions (e.g., favoring early or late options). In this section, we analyze the distribution of selections across the 10 positions for each cultural cluster, aggregated over all 17 models and 2,182 dilemmas (total valid selections: 36,977; expected per position under uniformity: 3,697.7). We quantify positional bias using Cramér's $V$ per cluster, interpreting values below 0.10 as negligible. Figure~\ref{fig:cluster_position_bias} shows selection percentages across display positions.

\paragraph{Overall Positional Bias Levels.}
All 10 clusters exhibit negligible positional bias, with Cramér's $V$ values ranging from 0.0510 (Nordic Europe) to 0.0877 (Eastern Europe), confirming no small, medium, or large effects across the board. This low bias underscores the effectiveness of our Latin Square design in mitigating ordering artifacts. A detailed breakdown can be found on GitHub.

\paragraph{Mild Primacy Effect in Positions.}
A modest yet discernible primacy tendency surfaces: Position 1 (P1) is chosen disproportionately in nine of the ten cultural clusters, 15.6\% for Confucian Asia and 15.5\% for Eastern Europe, for example, producing an aggregate deviation of 42.2 from the uniform baseline. Nordic Europe constitutes the sole exception, exhibiting a slight preference for P4 (12.7\%) over P1 (11.4\%). Choices in the later positions (P6–P10) remain evenly distributed, whereas P5 proves most balanced overall, with a deviation of only 8.6.

\paragraph{Cluster-Specific Variations.}
Clusters generally under-selected, Eastern Europe (2,086 total selections) and the Middle East (2,153), manifest the most significant positional biases ($V=0.0877$ and $V=0.0797$, respectively). Both clusters show elevated selection rates in early slots (Eastern Europe’s P1 at 15.5\%) and marked declines in later ones (P8 at 6.4\%). Conversely, highly favored clusters, such as Nordic Europe, register the lowest positional variance (2.34), indicating a more stable distribution across positions.

Together, these patterns confirm that the experimental protocol largely suppresses ordering effects; any residual positional bias is insufficient to compromise the cultural preference findings reported in the preceding RQs. The persistent over-selection of P1 suggests that future benchmarks could benefit from enhanced randomization schemes.

\subsection{Do LLMs Cluster by Cultural Behavior, and Does This Pattern Correlate with Their Geographic Provenance? (RQ5)}

To assess similarities in cultural selection patterns and their relation to developers' locations, we compute symmetrized Kullback-Leibler (KL) divergence between each pair of models' cluster distributions, yielding a 17$\times$17 distance matrix fed into average-linkage hierarchical clustering. Developer geographies are from public sources. Figure~\ref{fig:kl_divergence_heatmap} shows the KL-divergence heatmap and dendrogram.

\paragraph{Model Clustering Patterns.}
The dendrogram in Figure~\ref{fig:kl_divergence_heatmap}  reveals tight intra-family groupings at low divergence levels (e.g., Meta's Llama-3.3 and Llama-4 at 0.0033, Anthropic's Claude-3.7 and Claude-4 at 0.0398, Google's Gemini-2.0 and Gemini-2.5 at 0.0103, OpenAI's GPT-4o at 0.0200, Microsoft's Phi-4 and GPT-4.1 at 0.0032 via linkage). Broader clusters emerge around 0.01--0.05: one subgroup includes Google, OpenAI, xAI, Cohere, and Qwen; another features Anthropic, Meta, Perplexity, Mistral, and DeepSeek. 


\paragraph{Correlation with Geographic Provenance.}
Clustering shows weak geographic ties: US models span both subclusters, with corporate lineage (e.g., Meta/Anthropic vs. Google/OpenAI) dominating over location. Non-US models fragment similarly (Chinese: DeepSeek in subcluster 2, Qwen in 1; European-adjacent: Mistral in 2, Cohere in 1; AI21 with Google/Cohere). This implies that alignment strategies and shared corpora influence cultural behavior more than national origin, though Western-centric data limits distinctions.

\section{Discussion}
Our findings reveal a clear cultural bias in contemporary LLMs: when presented with plural legitimate options in a decision-making task, models tend to gravitate toward Nordic/Germanic clusters. The associated rationals are strikingly uniform: models rely on \textit{Future Orientation} and \textit{Performance Orientation} with regularity, while rarely grounding decisions in \textit{Assertiveness} or \textit{Gender Egalitarianism} (each $<3\%$). This does not merely imply that models have bias; rather, the very \textit{style} of advice being normalised in highly used models embodies a conciliatory, progress-centred, achievement-first worldview, which is systematically misaligned with contexts that legitimate hierarchy or where an explicit appeal to gender fairness is the appropriate move. Therefore, occasionally, a default to consensus-seeking performance talk can result in advice that comes across as naive or disrespectful. Models should instead surface plural options, making trade-offs explicit and labeling the value frames that motivate each option, allowing users to decide which options are valid given their local norms.

A notable limitation is our exclusive use of English prompts, which may bias outputs toward Anglo-American values and obscure regional nuances \cite{zhong_cultural_2024}; future work should utilize multilingual prompts to distinguish between language and cultural effects. If LLMs are to mediate everyday social decision-making, alignment must treat \textit{conflicts among legitimate values} as a design target, so that a model helping someone write to their supervisor or address a family dispute can offer guidance that is not simply \textit{good}, but \textit{good for their values}.  

\section{Conclusion}

In this work, we introduce \textit{CCD-Bench}, a novel benchmark specifically designed to assess how contemporary LLMs navigate conflicts among legitimate cultural values. Evaluating 17 LLMs on 2,182 dilemmas, we observed a pronounced cultural trend toward Nordic and Germanic clusters. Despite high rates of multi-dimensional rationales, models relied on a narrow set of GLOBE dimensions, while rarely invoking Assertiveness or Gender Egalitarianism ($<$3\% each), suggesting an aversion to confrontational or gender-focused reasoning and indicating limited true pluralism in reasoning. Our results underscore the need for more nuanced approaches to cultural dynamics in LLMs if they are to operate credibly within an increasingly pluralistic world.

\section{Acknowledgments}
This work is supported in part by the NYUAD Center for Interdisciplinary Data Science \& AI, funded by Tamkeen under the NYUAD Research Institute Award CG016.
\bibliography{aaai2026}

\end{document}